%% file: saclocalplanner.tex
\documentclass[letterpaper, 10 pt, conference]{IEEEconf}  % Comment this line out if you need a4paper

\IEEEoverridecommandlockouts                              % This command is only needed if 
\usepackage{graphicx} % for pdf, bitmapped graphics files
\usepackage{amsmath}  
\usepackage{amssymb}  % assumes amsmath package installed
\usepackage{xcolor}
\usepackage{subfigure}
\graphicspath{{./figs/}}
\usepackage{booktabs}
\usepackage{multirow}

\newcommand{\cA}{{\cal A}}
\newcommand{\cS}{{\cal S}}
\DeclareMathOperator*{\E}{\mathbb{E}}

\title{\LARGE \bf
SACPlanner: Real-World Collision Avoidance with a Soft Actor Critic Local Planner and Polar State Representations
}

\author{
Khaled Nakhleh\space\space Minahil Raza\space\space Mack Tang\space\space Matthew Andrews
 Rinu Boney\space\space Ilija Had\v{z}i\'{c}\space\space\\ Jeongran Lee\space\space Atefeh Mohajeri~\space\space
Karina Palyutina\\
Nokia Bell Labs - Murray Hill NJ, Espoo  Finland \& Cambridge UK
\thanks{Work performed while K.~Nakhleh, M.~Raza and M.~Tang were summer interns from Texas A\&M, Åbo Akademi University and U.~Maryland respectively.}% <-this % stops a space
}

\begin{document}

\maketitle
\thispagestyle{empty}
\pagestyle{empty}

%%%%%%%%%%%%%%%%%%%%%%%%%%%%%%%%%%%%%%%%%%%%%%%%%%%%%%%%%%%%%%%%%%%%%%%%%%%%%%%%
\begin{abstract}
We study the training performance of ROS local planners based on Reinforcement Learning (RL), and the trajectories they produce on real-world robots. We show that recent enhancements to the Soft Actor Critic (SAC) algorithm such as RAD and DrQ achieve almost perfect training after only 10000 episodes. We also observe that on real-world robots the resulting {\em SACPlanner} is more reactive to obstacles than traditional ROS local planners such as DWA. 
\end{abstract}

%%%%%%%%%%%%%%%%%%%%%%%%%%%%%%%%%%%%%%%%%%%%%%%%%%%%%%%%%%%%%%%%%%%%%%%%%%%%%%%%
\input{intro}
\input{RLenv}

\input{previous}
\input{sac}

\input{training}
\input{experimentdesign}
\input{experimentresults}
\input{trajectory}

\section{Conclusions and Future Work}
% In this work, we have examined the performance of RL-based local planners for ground robots. We use polar costmaps and regularization on top of SAC algorithm to achieve success rates close to 100\% after only 10,000 episodes. Our SAC agent outperforms Cartesian SAC agents as well as other RL methods. In addition, we have done a detailed trajectory analysis to show how the resulting SACPlanner is more robust and responsive to obstacles as compared to non-RL algorithms. For future work we would like to improve the smoothness of SACPlanner when there are no unexpected obstacles, and we plan to develop a cooperative version of SACPlanner for when two robots are in close proximity. For future work, we would like to improve the smoothness of SACPlanner when there are no unexpected obstacles, and we plan to develop a cooperative version of SACPlanner for when two robots are in close proximity.
In this work, we have examined how training for RL-based local planners can be improved by using polar costmaps and regularization on top of the SAC algorithm to achieve success rates close to 100\% after only 10,000 episodes. In addition, we have done a detailed trajectory analysis to show how the resulting SACPlanner is more robust and more responsive to dynamic obstacles than non-RL algorithms. For future work, we would like to improve the smoothness of SACPlanner when there are no unexpected obstacles, and we plan to develop a cooperative version of SACPlanner for when two or more robots are in close proximity.

\bibliographystyle{IEEEtran}
\newpage
\bibliography{ref}

\end{document}

%% file: intro.tex
\section{Introduction}
We study the efficacy of Reinforcement Learning (RL) algorithms for obstacle avoidance and local planning in ROS-based robotics
systems. RL algorithms are able to learn optimal actions based on a current state and a reward function.
The purpose of the ROS local planner is to adhere to a global
path to the current robot goal while avoiding local obstacles (which
may be dynamic). The RL paradigm is attractive for
such a problem since the behavior of an RL agent does not have to be
explicitly programmed for every possible scenario. 
In the RL framework, we specify the reward function, state space, and permissible actions the robot can take. The goal is to obtain a near-optimal planning policy given sufficient training samples. RL agents can potentially exhibit more complex (and hence more responsive) 
behavior than traditional local planners such as the Dynamic Window
Approach (DWA) to Collision Avoidance~\cite{fox1997dynamic}. 
%(We provide definitions of these algorithms below.)

RL has recently seen many advances due to the emergence of Deep RL,
where the actions are chosen from a policy parametrized by a 
Deep Neural Network (DNN). One notable success of Deep RL is in
learning policies for game environments (e.g.\ Atari games) modeled as Markov Decision Processes (MDPs) and standardized as {\em OpenAI Gym} environments~\cite{MnihKSGAWR13}. As a result
of this success, multiple authors have examined how Deep RL can be applied to robot control~\cite{GuldenringGHJ20,patel2021dwa,KastnerML20,liu2020robot}.

However, these works raise a number of questions that we address in our study. First, they typically measure performance via
an {\em episodic success} criterion, e.g.\ does the robot
reach the goal, does it suffer any collisions etc? We are also
interested in the {\em quality} of the trajectory. 
% We are also interested in the quality of the trajectory during our real-world experiments.
Is it smooth? How does it back off from an obstacle? Second, many of these papers address
challenging environments where success rates are significantly below
90\%. We believe such performance is unacceptable for practical
deployments. Therefore, we are interested in how to achieve near 100\%
success rates even in complex scenarios. % where that is achievable.  
Third, there are alternative obstacle-avoidance algorithms that do not use RL and we
would like to quantify the benefits and drawbacks of using an RL-based
approach. Lastly, we would like to know which specific RL techniques
produce the best performance.

%We focus on a cloud-based ROS environment in which most ROS
%nodes are migrated to an edge cloud. 
%We assume a physical
%configuration in which a map is available for the permanent
%configuration but other dynamic obstacles may also be present. We
%consider a single robot that must navigate to a sequence of goals
%without hitting any obstacles. When a goal is specified, the ROS
%{\em global planner} calculates a global route from the start position to
%the goal. The {\em local planner} is then responsible for following %the global plan without hitting any obstacles.
We follow \cite{guldenring2019applying,GuldenringGHJ20} and use ROS together with a {\em waypoint generator} that specifies a {\em next} waypoint based on the current robot location and a global plan to the goal. The task of the RL local planner is to reach this next waypoint without hitting any static or dynamic obstacles. Our RL state is an image representation of the obstacles and the next waypoint in polar coordinates. It mimics the image states used in the OpenAI gym environments for Atari games. 
%Our robot is equipped with a LiDAR scanner and
%the obstacle information is transmitted to the RL agent in the form
%of a costmap which we convert to polar coordinates. (An exact
%definition of this {\em polar costmap} is given in Section~\ref{}.)
%The trained RL agent then specifies a translational and rotational
%velocity pair $(v,\omega)$.  The RL state is an image formed from the
%polar costmap and the next waypoint. Since the state is an image we
%can utilize the RL frameworks that have been so effective for video
%games. (We refer to our RL state as a {\em gamelike state}.  We train
%within two environments. The first is a {\em dummyenv} that can
%produce polar costmaps from robot trajectories without the full ROS
%machinery. The second is a full-blown simulated ROS environment.
%With this setup we focus on the following questions.
We train our agents in a simulator with sample maps, and then upload the trained agents onto the robot for testing in the real world. With this setup, we list our contributions as follows:\\
$\bullet$ We show that modern variants of the Soft Actor-Critic (SAC) RL algorithm such as Reinforcement Learning with Augmented Data (RAD)~\cite{laskin2020reinforcement} and Data-regularized Q (DrQ)~\cite{kostrikov2020image} give significantly improved 
% training 
performance compared to earlier RL algorithms and implementations, and achieve success rates close to 100\% after only 10,000 episodes. We refer to the resulting local planner as {\em SACPlanner}.\\
$\bullet$ We demonstrate that polar image state representations outperform natural alternatives.\\
%We also observe that multi-stacking frames in the RL state does not improve performance.\\
$\bullet$ We analyze the trajectories produced by SACPlanner on real-world robots. (Prior work mostly limited trajectory analysis to simulations with perfect localization etc.) We compare with trajectories produced by DWA and a shortest-path based local planner. In all cases with an unexpected or dynamic obstacle, SACPlanner is much more reactive and hence performs better. The trade-off is a less smooth trajectory when the local planner simply has to follow the global plan.  
%$\bullet$ We observe that multi-stacking frames in the RL state leads to less responsive behavior by SACPlanner, even for dynamic obstacles. 
%$\bullet$ How does training with image-based gamelike states compare to other types of RL states that have been proposed in the literature?\\
%$\bullet$ How does training in an artificial environment like dummyenv compare to training directly in a ROS environment?\\
%$\bullet$ How can we improve training with modern variants of the Soft Actor Critic algorithm such as RAD~\cite{} and DrQ~\cite{}.\\
%$\bullet$ To what extent does multi-stacking frames improve the peformance with dynamic obstacles, both in terms of training and in terms of performance in real ROS environments?\\
%$\bullet$ What are the detailed dynamics produced by RL-based local planning algorithms? In particular, how noisy are they compared to traditional local planners such as DWA?\\
%$\bullet$ Can we replace an RL-based local planner with an algorithm based on repeated shortest paths?\\
%$\bullet$ What is the performance of RL-based local planners compared to traditional local planners like DWA? What are the scenarios where RL-based planners work well, and what are the scenarios where DWA works well? In particular, how do the two types of algorithms navigate different types of corners? How do they react to different types of obstacles?\\
%$\bullet$ How does moving to an RL-based local planner affect the interaction between the local and global planners?

%% file: RLenv.tex
\section{Training and Validation Framework} 
We use a standard ROS stack in which the robot knows its position up to the accuracy of the localization system. The robot has a 2D map for fixed, known obstacles and it detects dynamic and unknown static obstacles using a LiDAR sensor. From the raw obstacle information the robot constructs a costmap in the form of an Occupancy Grid using the approach of Lu et al.~\cite{lu2014layered}. The costmap window size for the local planner is $8m\times8m$.

We integrate RL into the robot navigation stack using the framework pioneered by G{\"u}ldenring et al.~\cite{guldenring2019applying,GuldenringGHJ20}. When a new goal is specified the global planner creates a path from the current position to the goal  (Fig.\ref{fig:ROS}). In this work we use without change the standard ROS {\em NavFn} planner based on the Dijkstra search algorithm. The path is found based on the obstacles in the map together with any obstacles seen by the LiDAR at the time of path creation.

Whenever a path is created by the global planner, a {\em waypoint generator} breaks it up into a sequence of waypoints. At all times the local planner maintains a list of 8 waypoints, starting with the one after the waypoint that is closest to the robot. (The method of \cite{guldenring2019applying} sometimes starts the list with the closest waypoint to the robot, but we found that could create excessive ``pingponging'' in the eventual choice of waypoint). From this list of 8, the local planner chooses the first on the list that is not too close to an obstacle. The aim of the RL agent is to move towards the selected waypoint while not hitting any obstacles, {\em including obstacles that appeared after the global plan was computed}.

\subsection{RL Environment} \label{section:Environments}

% RL problems are often formalized as Markov decision processes~(MDPs). An MDP is defined by a tuple $(\mathcal{S}, \mathcal{A}, p, R, \gamma)$, where the set of states $\mathcal{S}$ obey the Markov property $p(s_{t+1}| s_t ) = p(s_{t+1} | s_1,..,s_t )$, $\mathcal{A}$ is a set of actions, $p$ is the state-transition distribution $s_{t+1} \sim p(\cdot|s_t, a_t)$ that represents the probability of transitioning to state $s_{t+1}$ after taking action $a_t$ in state $s_t$ at any timestep $t$, $R(s_t,a_t)$ is the reward function, and $\gamma\in[0,1]$ is the discount factor which reduces the weight given to future rewards.

The RL environment is defined  by a state space $\cS$, an action space $\cA$, and a reward function $R(\cdot,\cdot)$. When the RL agent takes action $a \in \cA$ in state $s \in \cS$, it gains reward $R(s, a)$ and moves to a new state $s'$ according to some state-transition distribution $s' \sim p(\cdot|s, a)$.
The actions are linear/angular velocity pairs $(v,\omega)$. The state space is defined by the positions of the next waypoint and the local obstacles relative to the current robot position. We represent the state with an image since this allows us to utilize the convolutional deep RL architectures that have worked well for visually-rich environments such as Atari video games and some robot control tasks.
% (we defined such architectures in the next section). 
In addition, using such game-like image states is a convenient way to merge the information from the waypoint position, the static objects from the map, and the dynamic obstacles sensed by the LiDAR. %(note that most of the nearby map obstacles will also be sensed by the LiDAR).

\begin{figure}[!t]
	\centering
	\subfigure{\includegraphics[height=0.13\textwidth]{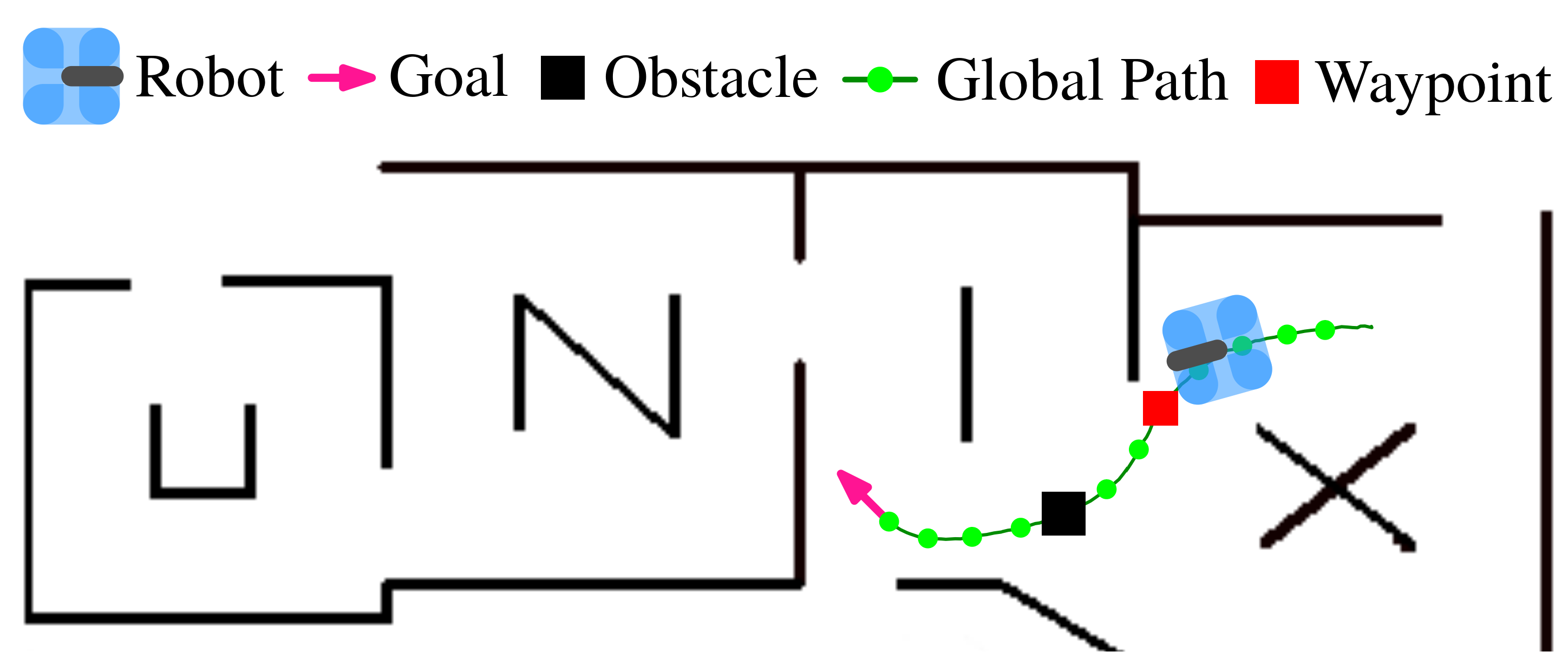}}\hspace{2mm}
	\subfigure{\includegraphics[height=0.13\textwidth]{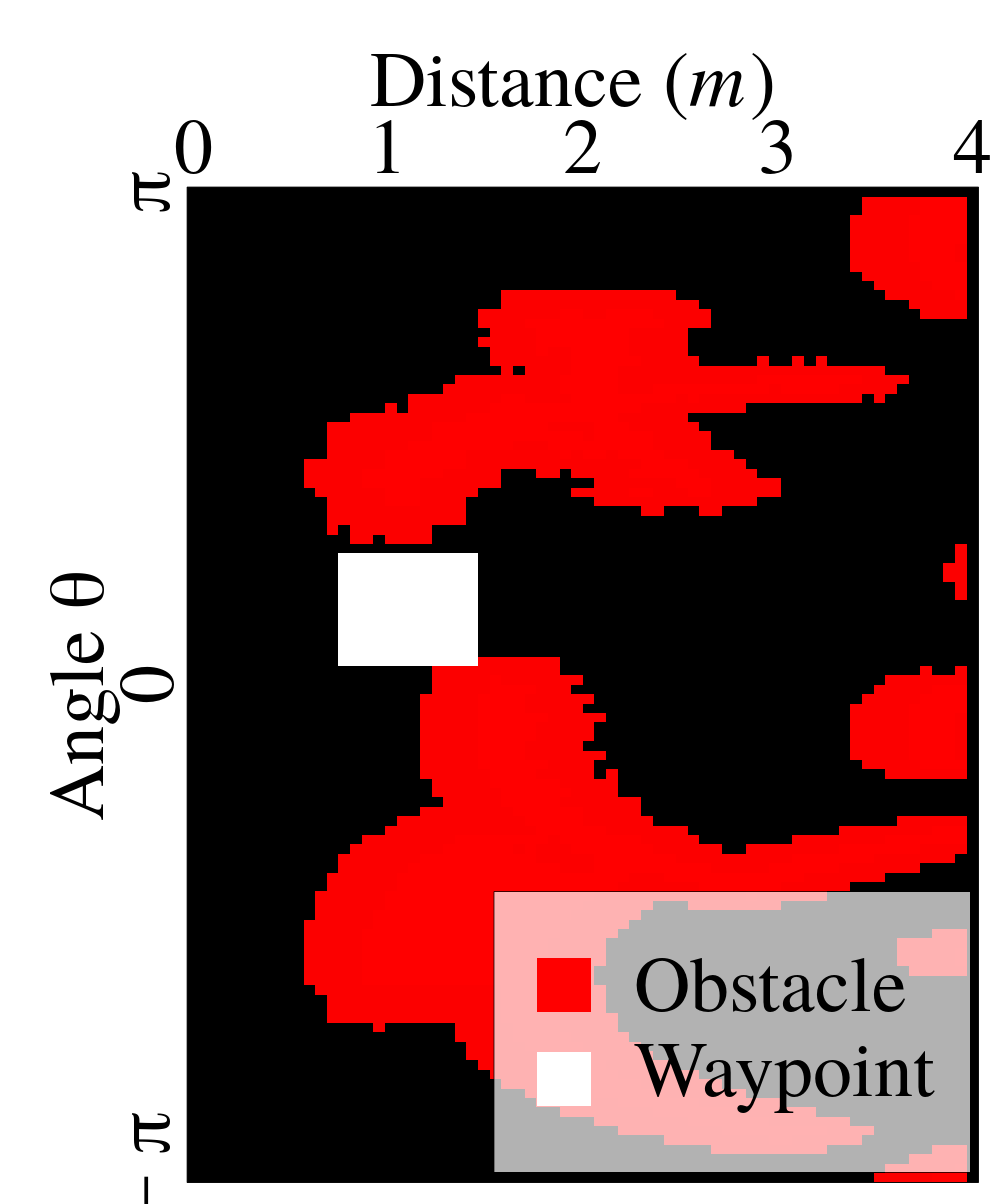}}
	\vspace{-1mm}
	\caption{ROS framework with global map, and polar costmap. The black square represents an obstacle that appeared after the global plan was computed.}\label{fig:ROS}
\end{figure}

%We have two ways to represent the state as an image (illustrated in Fig.\ref{fig:ROS}), both of which are derived from the ROS costmap. The first, that we refer to as the {\em Cartesian costmap} simply replicates the occupancy grid representation of the costmap with red regions denoting the obstacles. The robot is assumed to be at the center of the costmap and the next waypoint is represented with a white square. 
%{\sc Need to define the orientation.} 
Specifically, our RL state is an image that we refer to as the {\em polar costmap}. (See Fig.~\ref{fig:ROS}.) It is generated by converting the Occupancy Grid representation of the ROS costmap and the next waypoint to polar coordinates. The horizontal axis represents distance from the robot and the vertical axis represents angle. Obstacles are presented in red and the next waypoint is a white square. 
% as in Fig.\ref{fig:cart_vs_polar}. 
The motivation for using a polar representation is that it matches the linear/angular velocities that form the action. The state transition naturally follows from the robot movement after an action is taken.

It remains to define the reward function $R(s,a)$ for taking action $a$ in state $s$. We employ a mix of both dense and sparse rewards. For a given state $s$, let $(d_{\mathrm{old}}, \theta_{\mathrm{old}})$ be the distance and bearing
to the next waypoint in state $s$, let $s'$ be the new state after taking action $a$, 
and let $(d_{\mathrm{new}}$, $\theta_{\mathrm{new}})$
be the distance and bearing in state $s'$. Here the bearing is defined to be the difference between the angle to the waypoint and the current yaw. 
We define:
$$
\begin{array}{rl}
    & R(s,a)=\left(d_{\mathrm{old}}- d_{\mathrm{new}}\right) \cdot\left(1 \mbox{~if } d_{\mathrm{old}}- d_{\mathrm{new}} \ge 0, \mbox{~else } 2 \right) \\
    &+\left(|\theta_{\mathrm{old}}|-|\theta_{\mathrm{new}}|\right)\cdot\left(1 \mbox{~if } |\theta_{\mathrm{old}}|-|\theta_{\mathrm{new}}| \ge 0, \mbox{~else } 2 \right) \\
    &-R_\mathrm{max} \cdot\left( 1 \mbox{~if collision, else } 0 \right)  \\
    &+R_\mathrm{max} \cdot\left( 1 \mbox{~if } d_{\mathrm{new}}=0, \mbox{~else } 0\right)\\&-G(s'),
  \label{eq:rew}
\end{array}
$$
where $R_\mathrm{max}$ is a fixed reward/penalty for reaching the waypoint and colliding with an obstacle, respectively, and $G(s')$ is the product of a truncated Gaussian kernel centered at the robot location and the Occupancy Grid in state $s'$. 

The first two terms of $R(s,a)$ incentivize getting closer to the waypoint both in terms of distance and bearing. Note that the penalty for moving away from the waypoint (both in distance and bearing) is double the reward for moving towards it. Hence there is a net penalty for moving away from the waypoint and then back towards it. We have found that this ``doubling the penalty for negative progress'' has a significant effect on encouraging the agent to move directly to the waypoint if there are no obstacles in the way. The final Gaussian term penalizes movement towards an obstacle. 

\begin{figure}[h]
	\centering
	\subfigure{\label{fig:dummy_cart}
	\includegraphics[height=0.12\textwidth]{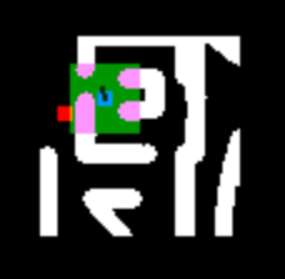}}%\hspace{-1mm}
	\subfigure{\label{fig:dummy_polar}
	\includegraphics[height=0.12\textwidth]{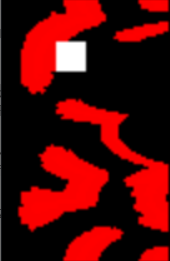}}
	\vspace{-2mm}
	\caption{Dummy training environment (left) with polar costmap (right).}\label{fig:dummy}	
\end{figure}

We find that it is more efficient to train our RL agents on a ``dummy'' training environment that does not require the full complexity of ROS or a detailed physics simulation. For this dummy training environment we place a robot start position and a single waypoint in an environment with obstacles as shown in Fig.\ref{fig:dummy}. The robot is the blue square, the waypoint is the red square, and the larger green square around the robot is the support of the truncated Gaussian kernel. For each episode in the RL training, we pick an obstacle configuration and then use the above reward to encourage the RL agent to move towards the waypoint without hitting obstacles. %For some episodes all obstacles are static and for others we also move dynamic obstacles during the training. 
Once the agent is trained we can run it directly in our ROS environment (either a Gazebo simulation or on real robots) since the state definition is the same in all cases. We remark, however, that the specific obstacle configurations on which we do the training are {\em not} the same as the configurations on which we do our eventual experiments, since we want trained agents that generalize to any unseen obstacle configuration.

%% file: previous.tex
\section{Previous Work and Comparison Algorithms}
\label{sec:previous}
The canonical local planner algorithm for ROS is the Dynamic Window Approach~\cite{fox1997dynamic}. At each instant, DWA calculates a set of achievable $(v,\omega)$ pairs based on the current velocities and achievable acceleration characteristics of the robot. For each velocity pair 
%(which corresponds to a circular arc), 
DWA calculates a score based on how closely the arc follows the global plan, and on how far the arc is from any obstacle. It then chooses the best velocity pair based on this score. 

Multiple recent papers have investigated how well local planner behavior can be learned via RL. G\"uldenring et al.~\cite{guldenring2019applying} developed a framework that has been followed by many subsequent papers in which the global plan is partitioned into waypoints and the task of the RL agent is to get to the next waypoint. 
%Our work also follows Guldenring's basic approach but we extend with polar-based image states and also investigate more recent RL algorithms. 

Patel et al.~\cite{patel2021dwa} combines the DWA and RL approaches. The resulting DWA-RL algorithm calculates a cost for each potential velocity pair, but then uses RL to select the best pair based on the full spectrum of costs, rather than just picking the lowest cost pair. The work of K\"astner et al.~\cite{KastnerML20} distinguishes between humans, robots and static objects and uses an RL state that is a combination of the raw LiDAR input, the distance/angle to the goal, the position of nearby humans and the position of nearby robots. A follow-up paper~\cite{KastnerZBLSLM21} looks at different methods for choosing the next waypoint, and compares the fixed partition of  G\"uldenring et al.~\cite{guldenring2019applying,GuldenringGHJ20}, with alternative methods that choose the waypoint more dynamically. The work of Liu et al.~\cite{liu2020robot} uses a similar RL state. The main difference is that they represent pedestrian and robot movement using the CrowdNav algorithm of \cite{chen2019crowd}, and they represent the LiDAR information via both the raw LiDAR values and an Occupancy Grid.

In many of these papers the success rate of the trained agent is significantly under 100\%. For example, the agent of \cite{guldenring2019applying} converges to a rate less than 70\%. Moreover, this prior work typically provides trajectory plots from a Gazebo simulation. Our goal is to train an agent with close to 100\% success, and then analyze trajectories from a real-world deployment (with the associated imperfections in sensing and localization). We also observe that if the goal is to get to the next waypoint, then an alternative is to repeatedly calculate a shortest path in the Occupancy Grid. We have found that modern python implementations of Dijkstra's algorithm can do this sufficiently fast, and so we also compare against a local planner that uses the next segment of the shortest path to define the robot velocities. Note however that the shortest path will change over time as the robot and obstacles move.  

%% file: sac.tex
\section{Soft Actor Critic Algorithm}
% start with actor-critic?
% What is the reason for using soft actor critic? soft value function
% table summarizing refinements?
%Reinforcement learning (RL) aims to learn intelligent behaviors through interactions with the environment.
%The policy function $a \sim \pi(\cdot|s)$ of an RL agent is a mapping from states to actions and defines the behavior of the agent.
The objective in RL is to maximize the expected sum of rewards that the agent will receive in the future: $G = \E [ \sum_{t=0}^\infty\gamma^t R(s_t, a_t) ]$,
%\begin{align*}
%    G = \E \left[ \sum_{t=0}^\infty\gamma^t R(s_t, a_t) \right] \,,
%\end{align*}
where the expectation is taken over the agent policy $a_t \sim \pi(\cdot|s_t)$ and the state transition function $s_{t+1} \sim p(\cdot|s_t, a_t)$. The parameter $\gamma \in (0,1]$ is a discount factor used to reduce the weight given to future rewards.
% $a_t$ is the action taken by the agent at timestep $t$, $s_{t+1} \sim p(\cdot|s_t, a_t)$ is the state transition function, and $\gamma \in [0,1]$ is a discount factor which reduces the weight given to future rewards. 

%The long-term reward (return) $G_t$ is given by the equation \ref{eq:return}
%\begin{equation}\label{eq:return}
%    G_t = \Sigma_{k=0}^\infty\gamma^k\ r_{t+k+1}
%\end{equation}
%is used to calculate the long-term reward (return) at a given time-step $t$.

Continuous control problems, such as the local navigation task considered in this paper, are often approached using actor-critic algorithms that learn two functions called the actor and the critic. The actor is a policy function $a \sim \pi_\theta(\cdot|s)$ with parameters $\theta$. % defines the behavior of the agent. 
The critic $Q_\phi(s, a)$ with parameters $\phi$ estimates the action-value function 
% $Q^\pi(s, a)$ 
$Q^\pi(s, a) = \E [ \sum_{t=0}^\infty\gamma^t R(s_t, a_t) | s_0 = s, a_0 = a ]$
of policy $\pi$, which is the expected cumulative reward after taking action $a$ in state $s$ and following policy $\pi$ after that.
% \begin{align*}
%     Q^\pi(s, a) = \E \left[ \sum_{t=0}^\infty\gamma^t R(s_t, a_t) \bigg| s_0 = s, a_0 = a \right] \,.
% \end{align*}
% Actor-critic algorithms are a class of RL algorithms which consist of two networks- the actor and the critic. 
% The actor observes the state of the environment and outputs an action while the critic evaluates how good the action was \cite{actorcritic}. 

In this work, we use a state-of-the-art off-policy actor-critic algorithm called Soft Actor-Critic (SAC) \cite{haarnoja2018softa, haarnoja2018softb}. It is based on the maximum entropy RL framework which augments the standard RL objective with an entropy maximization objective: $G = \E [ \sum_{t=0}^\infty\gamma^t ( R(s_t, a_t) + \alpha \mathcal{H}(\pi(\cdot|s_t))) ]$,
% \begin{align*}
%     G = \E \left[ \sum_{t=0}^\infty\gamma^t \Big( R(s_t, a_t) + \alpha \mathcal{H}(\pi(\cdot|s_t)) \Big) \right] \,,
% \end{align*}
where $\alpha$ is a learnable temperature parameter that balances the importance of both objectives. The entropy maximization motivates the agent to succeed at the task while acting as randomly as possible, aiding exploration.
% which aims to maximize the entropy in addition to the standard RL objective. We want the RL agent to succeed at the task while acting as randomly as possible which incentivizes the policy to explore more widely.

% A policy $\pi(a|s)$ is a probability distribution over the set of actions $A$ given the current state $s\in S$. We use a neural network to learn the mean and standard deviation of the policy. The action-value function, also referred to as the Q-function $Q^\pi(s,a)$ is the expected return when starting in a state $s$, taking the action $a$ and following $\pi$. In this work, the “critic” estimates the two Q-functions while the “actor” updates a Gaussian policy in the direction suggested by the critic. %Note that we take a minimum of the two Q values. 
%The state-value function $V^\pi(s)$ for a policy $\pi$ is the expected return when starting in a state $s$ and following $\pi$.of taking action $a$ in state $s$ under $\pi$ is the expected return when starting in a state $s$, taking the action $a$ and following $\pi$. In the actor-critic framework, the “critic” estimates the value function (the Q value or the V value) while the “actor” updates the policy in the direction suggested by the critic.
% \begin{equation}
%     Q(s,a) = \mathbb{E}_\pi [G_t | s_t=s, a_t=a]
% \end{equation}

In SAC, the actor and critic functions are parameterized as deep neural networks. The actor is a Gaussian policy with the mean and diagonal covariance parameters produced by the neural network.
% The policy and twin Q functions, parameterized by $\theta$, $\phi_1$ and $\phi_2$ respectively, 
The actor and critic networks are updated by sampling minibatches of $(s_t, a_t, r_t, s_{t+1}, d_t)$ transitions 
%$\mathbb{B}$ 
from a replay buffer $\mathcal{D}$, where $d_t$ is a terminal signal denoting the end of the episode.
% The batch of transitions consist of the current state $s$, the action $a$, the next state $s_t+1$, the reward $R$ and the done signal $d$ represented by the tuple $(s, a, R, s_{t+1}, d)$. 
The parameters for the critic network $Q_\phi$ are trained to minimize the soft Bellman residual:
\begin{equation*}
    J_Q(\phi) = \mathbb{E}_{(s_t, a_t, r_t, s_{t+1}, d_t) \sim \mathcal{D}} \left[ Q_\phi(s_t, a_t) - y_t \right]^2 \,,
\end{equation*}
where the learning target $y_t$ is
\begin{equation*}
y_t = r_t + \gamma (1 - d_t) V(s_{t+1}) \,,
\end{equation*}
and the soft value function
\begin{equation}
V(s_t) = E_{a_t \sim \pi(\cdot|s_t)} \left[ Q_{\bar{\phi}}(s_t, a_t) - \alpha \log \pi(a_t|s_t) \right]
\label{eq:v}
\end{equation}
is approximated using a Monte Carlo estimate of the policy $\pi_\theta$ and a target Q network $Q_{\bar{\phi}}(s_t, a_t)$ whose parameters $\bar{\phi}$ is maintained as the exponentially moving average of the Q network parameters $\phi$.
SAC also makes use of clipped double Q-learning \cite{fujimoto2018addressing}, where the Q estimates are computed as the minimum value of an ensemble of two critic networks with different initializations trained on the same data. This helps prevent overestimation bias in Q-learning with non-linear function approximators.

% The target for the Q function $y$ is defined as
% \begin{multline}
%     y(R, s_{t+1}) = R + \gamma(1-d)(\min(Q_\phi_i(s_{t+1},a_{\theta}))-\\ \alpha \log \pi(a_{t+1}|s_{t+1})), a_{\theta} \sim \pi(.|s_{t+1})
% \end{multline}
%value function $V_\psi^-$ is implicitly parameterized through the soft Q-function $Q_\phi^-$ which has been shown to stabilize training. 
% \begin{equation}
%     V(s_t) = \mathbb{E}_{a_t\sim\pi}[Q_\phi(s_t, a_t)- \alpha log \pi(a_t|s_t)] 
% \end{equation}
% Finally, the policy is learned by minimizing the divergence from the exponential of the soft-Q function at the same states
% \begin{equation}
% J_\pi(\theta)=- \frac{1}{\mathbb{B}} \sum_{s\in\mathbb{B}}[\min(Q_\phi_i(s,a_{\theta})-\alpha \log \pi_\theta (a_{\theta}|s_t)]    
% \end{equation}

The parameters of the actor/policy network $\pi_\theta$ are updated to maximize the maximum entropy RL objective:
\begin{equation*}
J_\pi(\theta) = \E_{s_t \sim \mathcal{D}} \left[ \E_{a_t \sim \pi_\theta(\cdot|s_t)} \left[ \alpha \log \pi_\theta(a_t|s_t) - Q_\phi(s_t, a_t) \right] \right] \,.
\end{equation*}

The learnable temperature parameter $\alpha$ can be automatically updated such that the policy network satisfies a minimum expected entropy constraint. See \cite{haarnoja2018softb} for more details.

While SAC often performs well on continuous control tasks with low-dimensional observations, learning a mapping from high-dimensional states (images) to continuous actions (linear and angular velocities) typically requires massive amounts of robot-environment interactions.
%high dimensional inputs such as images is challenging. 
This is because the agent must learn to extract the right information from the images to successfully perform the task at hand. SAC with a convolutional encoder can be used to learn low-dimensional representations of image observations, which are then provided to the actor and critic networks. However, this often fails. Sample-efficient learning of SAC agents from image observations requires additional supervision such as input reconstruction~\cite{yarats2021improving}, contrastive representation learning~\cite{srinivas2020curl}, or image augmentations~\cite{laskin2020reinforcement, kostrikov2020image}. 

In this work, we consider the recently proposed RAD~\cite{laskin2020reinforcement} and DrQ~\cite{kostrikov2020image} methods that apply image augmentations for sample-efficient learning of continuous control policies from image observations. 
% Both RAD and DrQ propose to augment the image observations by randomly shifting the images by a few pixels. 
In RAD and DrQ, the image observations are transformed with a random shift before each forward pass on the convolutional encoder. DrQ further proposes to average the Q-learning targets in Eq.~\ref{eq:v} over $K$ image transformations. This reduces the variance in the learning targets of the critic, improving the stability and efficiency of learning.

% Real-world implementations require months of training. Therefore, optimizing performance with limited environment interaction is pivotal to the applicability of RL in real world scenarios. 
% To exploit the structure of MDP, we use optimality invariant state transformation suggested in CITEDRQ. An optimality invariant state transformation $\mathbb{f}: \mathbb{S}\times\mathbb{T}\rightarrow\mathbb{v}$ is a mapping that preserves Q values as shown in (\ref{eq:Q_tranform}). An example of
% is random image translation (RandomCrop).
% \begin{equation}\label{eq:Q_transform}
%     Q(s,a) = Q(\mathbb{v}, a)
% \end{equation}

% This allows the creation of K surrogate states by applying transformations to each state sampled from the replay buffer, $s\sim\mathbb{D}$ at every update step. We average the target values over K transformations which reduces the variance of the Q function. 
% \begin{multline}
%     \mathbb{y}(R,s_{t+1},d)=R + \frac{\gamma(1-d)}{K}\\\Sigma_{k\sim K}(\min(Q_\phi_i(s_{t+1}^k,a_{\theta}^k))- \alpha \log \pi(a_{t+1}|s_{t+1}^k)) 
% \end{multline}
% Additionally, we generate M transformations to compute the Q function values used for calculating the critic loss.
% \begin{multline}
%     J_Q(\psi)= \frac{1}{\mathbb{B}}\Sigma_{m\sim M,\mathbb{B}} (Q_\phi_i(m,a) - \mathbb{y}(R, s_{t+1}^m, d))^2
% \end{multline}
% For K=1 and M=1, these augmentations reduce to image transformations proposed in CITERAD.

% Summary of the RandomCrop Augmentation
We apply random shift image augmentation (by $\pm4$ pixels) to the costmap observations.
% in our local navigation task. For example, the polar costmap images are $116 \times 68$. We pad each side by repeating the bounding pixels to get an image of size $120 \times 72$. Then we randomly crop the image to the original size $116 \times 68$, yielding the original image shifted by $\pm4$ pixels. 
% summary of encoder architecture
The augmented images are passed to a convolutional encoder consisting of four $3\times3$ convolutional layers with 32 filters and a stride of 1 followed by ReLU activation. The output of the final convolutional layer is flattened and passed to a fully connected layer, followed by layer normalization and tanh activation to yield a 50-dimensional state representation. 
%summary of policy and critic network
The actor and critic networks use the same MLP architecture with 4 fully connected layers of 1024 hidden units. The actor predicts the mean and diagonal covariance of the Gaussian policy based on the encoded state vector. The critic networks predict the scalar state-action values based on the encoded state vector and an action vector.
% Encoder and critic target networks
% Lastly, we use target networks for the encoder and the critic weights to  stabilize the training
Following previous works \cite{yarats2021improving, srinivas2020curl, laskin2020reinforcement, kostrikov2020image}, we train the convolutional encoder network using only the critic loss and then detach the network parameters from the actor loss for improved training stability.

%We use image augmentation proposed in CITERAD to learn directly from images without the need for learning simplified representations separately. For greater sample efficiency, we use the work in CITEDRQ which introduces a framework for regularizing the value function through transformations of the input state. These transformations create surrogate states to reduce the variance of Q-function. Two regularization methods have been combined by averaging the Q target over K image transformations and averaging the Q function over M image transformations. The averaged values are used to compute the critic loss for learning the Q-function.

% image is high dimensional. encoder produces a 50 feature output. 

%{'At goal': 986, 'Episode timeout': 11, 'Collision': 3} 40.96621973674467
% 9 impossible situations
%{'At goal': 986, 'Episode timeout': 8, 'Collision': 2} 40.96621973674467

% {'At goal': 978, 'Collision': 9, 'Episode timeout': 11, 'Lost sight of the goal': 2} 40.05592110671654
% {'Collision': 487, 'At goal': 417, 'Lost sight of the goal': 78, 'Episode timeout': 18} -6.293488300160666

%% file: training.tex
\section{Performance of RL Training}
\label{sec:training}

We now evaluate the training performance of RAD, DrQ and other baseline RL methods in our dummy training environment. We train the RL agents on our polar costmap environment from  Section \ref{section:Environments}, and compare against Cartesian costmap environments (similar to \cite{guldenring2019applying}) where we do not convert to polar coordinates before generating the image. We train for 10,000 episodes with the hyper-parameter values listed in Table \ref{tab:hyperparameters}, %using an RTX A4000 Nvidia GPU --> all models are not trained on this GPU 
% comparison of polar and cartesian costmap behaviour
% for cartesian costmaps, the agent overshoots the goal and loses sight of the goal
The trained agents are evaluated over 1000 episodes in the training environment. We define the success rate as the percentage of episodes in which the agent reached the goal. The collision rate is defined as the percentage of episodes where the agent collides with the obstacles. An episode can be neither a success nor a collision if the robot stops and the episode times out. % should we mention that we exclude impossible scenarios?
\begin{table}[t!]
    \centering
    \caption{Hyper-parameter values for SAC agent training.}
    \begin{tabular}{l|c}
        \toprule
        Hyper-parameter & Value \\
        \midrule
        Training episodes &  10000\\
        Random exploration episodes & 10\\
        %Max. steps per episode & 200\\
        Mini-batch size & 128\\
        Replay buffer capacity & $10^6$\\
        Discount factor $\gamma$ & 0.99\\
        Optimizer & Adam\\
        Learning rate & $0.001$\\
        Critic target update frequency & 2\\
        Critic target update rate $\tau$ & $0.01$\\
        Actor update frequency & 2\\
        \bottomrule
    \end{tabular}
    
    \label{tab:hyperparameters}
\end{table}

\begin{table}[t]
	\caption{Comparison of Cartesian and polar costmaps using RAD agent in the dummy environment.}
	\label{tab:cartesian_vs_polar}
    \centering
	\begin{tabular}{l|l|r|r}
	\toprule
	\multirow{2}{*}{Costmap} & Orientation & Success & Collision \\
	& Information & Rate & Rate\\
	\midrule
	Polar & Implicit & 98.7\% & 0.08\% \\
	\midrule
	\multirow{3}{*}{Cartesian} & Rotation & 42.0\% & 37.7\% \\
	& Arrow & 45.5\% & 36.0\% \\
	& Channel & 65.3\% &25.9\% \\
	\bottomrule
	\end{tabular}
\end{table}

\begin{table}[t]
	\caption{Comparison of RL agents in the dummy environment.}
	\label{tab:rl_agents}
    \centering
	\begin{tabular}{l|r|r}
	\toprule
	Method & Success Rate & Collision Rate \\
	\midrule
    %DQN & 46.3\% & 4.74\% \\ %500M steps
    DQN & 33.9\% & 51.2\% \\
    PPO & 83.6\% & 7.5\% \\
    SAC from LiDAR & 34.2\% & 47.5\% \\
    DWA-RL with SAC & 7.3\%  & 70.3\% \\
	RAD & 98.7\% & 0.08\% \\
    \textbf{DrQ} & \textbf{99.4\%} & \textbf{0.02\%} \\
	\bottomrule
	\end{tabular}
\end{table}

In Table~\ref{tab:cartesian_vs_polar}, we compare the polar and Cartesian costmaps using the RAD version of SAC.
While the information regarding the robot's orientation is implicit in the polar costmap, this information is missing in the Cartesian costmap. 
We explored three ways to represent this: (i) rotating the Cartesian costmap by the robot orientation angle, (ii) drawing an arrow at the center of the costmap to denote the robot orientation, or (iii) appending an extra channel to the costmap with the robot orientation angle.
% conveyed using rotation of the local costmap, drawing an arrow at the center of the costmap and by appending an extra channel to the image containing the robot's orientation.
The agent with polar costmap observations significantly outperforms those with Cartesian costmap observations. We hypothesize that this is because the polar costmaps better match the action space of the robot and also implicitly represent the robot orientation information, which allows for better generalization.
%We observed that RAD agents trained using the Cartesian costmap tend to overshoot the goal, which results in an episode timeout or collision in the majority of the episodes. 
We use the better performing polar costmap environment in the rest of our experiments.

We next compare the performance of RAD, DrQ (with K=2), and the following RL baselines in Table~\ref{tab:rl_agents}:

\noindent $\bullet$ \textbf{DQN}. To evaluate if discrete control is easier to learn, we discretize the action space of the robot with six possible linear/angular velocity pair combinations and train a standard DQN agent from the stable baselines library \cite{stable-baselines3}.
%on this discretized dummy environment.

\noindent $\bullet$ \textbf{PPO}. To evaluate if the SAC agents perform better than other actor-critic algorithms, we also compare against the popular PPO agent from the stable baselines library \cite{stable-baselines3}.

\noindent $\bullet$ \textbf{SAC from raw LiDAR observations}. To evaluate the importance of image-based game-like states, we compare against a SAC agent trained on raw LiDAR observations (similar to \cite{KastnerML20,liu2020robot}). For this agent, the actor and critic networks receive state vectors consisting of the raw LiDAR readings and the coordinates of the next waypoint. 

\noindent $\bullet$ \textbf{DWA-RL with SAC}. To evaluate if it is beneficial to combine the standard DWA planner with RL, we implement the observation space and reward function of the DWA-RL method \cite{patel2021dwa} and train our SAC agent on this hybrid setup.

% Table~\ref{tab:dummyenv_results} compares the polar SAC agents trained with RAD and DrQ (with K=2 augmentations per observation) agents. 
The DrQ method achieves the highest success rate ($> 99\%$) with the fewest collisions. 
We also experimented with stacking four consecutive frames as observations to the DrQ method but observed that these agents tend to have trouble navigating around obstacles, reducing the success rate to $94.9\%$.
We note that the success rates we obtain with the baseline algorithms are lower than those observed in the literature~\cite{guldenring2019applying,patel2021dwa,KastnerML20,liu2020robot}. We believe this is partly because we only run for 10,000 episodes (which corresponds to $<500000$ steps). However, this is sufficient for training the DrQ agent and demonstrates the sample-efficiency of this variant of SAC. Another potential reason is that our training environment contains challenging scenarios requiring tight turns (see Fig.~\ref{fig:dummy}), but this is necessary to obtain agents that will work for the real-world cases described below. 

% stable baselines comparisons TODO
% Rinu: remove Average reward to fit table into column?
% \begin{table}[t!]
% 	\caption{Comparison of RL Agents in the Dummy Environment}
% 	\label{tab:dummyenv_results}
% 	\begin{tabular}{l|l|l|r|r|r}
% 	\hline
	
% 	Costmap&Method&Orientation&Success&Collision& Average\\
% 	&&information&Rate &Rate&Reward\\\hline
% 	\multirow{2}{*}{Polar} & DrQ &\multirow{3}{*}{Implicit}	 & 99.4\% & 0.2\%&40.97 \\\cline{2-2}\cline{4-6}	
% 	 &  RAD & & 98.7\% & 0.8\%&40.01 \\\cline{1-2}\cline{4-6}
% 	 Polar (stacked)& DrQ &  & 94.9\% & 0.06\%&39.32 \\\cline{1-6}
% 	\multirow{3}{*}{Cartesian} & RAD & Rotation & 42.0\% & 37.7\% & -6.29 \\\cline{2-6}
% 	 & RAD & Arrow & 45.5\% & 36.0\%&-4.11 \\\cline{2-6}
% 	 & RAD & Channel & 65.3\% &25.9\%&13.39 \\\cline{1-6}
% 	\end{tabular}
% \end{table}

% \begin{table}[t!]
% 	\caption{Comparison of RL Agents in the Dummy Environment}
% 	\label{tab:dummyenv_results}
% 	\centering
% 	\begin{tabular}{|l|l|l|r|r|}
% 	\hline
% 	Costmap&Method&Orientation&Success&Collision\\
% 	&&information&Rate &Rate\\\hline
% 	\multirow{2}{*}{Polar} & DrQ &\multirow{3}{*}{Implicit}	 & 99.4\% & 0.02\%\\\cline{2-2}\cline{4-5}	
% 	 &  RAD & & 98.7\% & 0.08\% \\\cline{1-2}\cline{4-5}
% 	 Polar (stacked)& DrQ &  & 94.9\% & 0.06\% \\\cline{1-5}
% 	\multirow{3}{*}{Cartesian} & RAD & Rotation & 42.0\% & 37.7\% \\\cline{2-5}
% 	 & RAD & Arrow & 45.5\% & 36.0\% \\\cline{2-5}
% 	 & RAD & Channel & 65.3\% &25.9\% \\\cline{1-5}
% 	\end{tabular}
% \end{table}

%% file: experimentdesign.tex
\section{Design of Robot Experiments}
We now describe our experiments for testing the local planners on a physical robot. We use a ClearPath Robotics Jackal robot~\cite{JackalDatasheet} equipped with LiDAR, set to a scanning frequency of 5Hz. 
The experiments cover a range of scenarios that an autonomous robot would encounter in the physical world.
%and is required to traverse successfully.
A failed traversal translates into a robot's collision with a static obstacle (e.g. wall), or a dynamic obstacle (e.g. pedestrian).
Moreover, if the planner fails to complete the global plan, then the robot fails that scenario. In addition to simply measuring success/failure, we are also interested in the nature of the trajectory produced by each approach. Is it smooth? How does the robot react to an obstacle? 

\textbf{Test cases.} The experiments were conducted in a facility that includes an open room and a maze component with tight corners and narrow doorways shown in Fig.\ref{fig:test_cases}. We refer to the maze shown in the first two images of Fig.\ref{fig:test_cases} as the \emph{UNIX maze room} (named after letters that make up obstacles in four separate rooms). We describe four test cases:

\begin{figure}[!t]
	\centering
	\subfigure{\label{fig:test1}
		\includegraphics[width=0.15\textwidth]{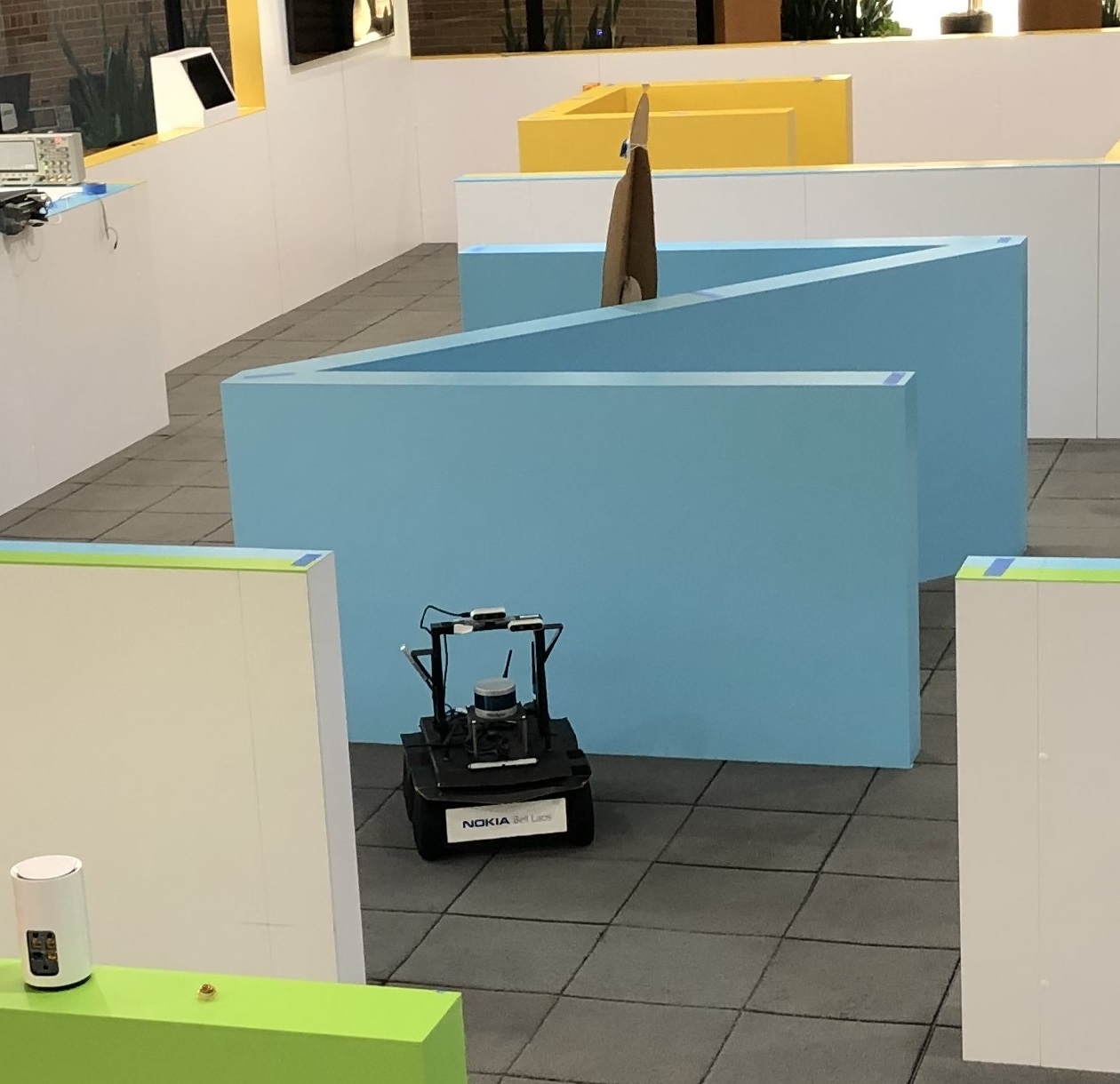}}\hspace{-1mm}
	\subfigure{\label{fig:test2}
		\includegraphics[width=0.15\textwidth]{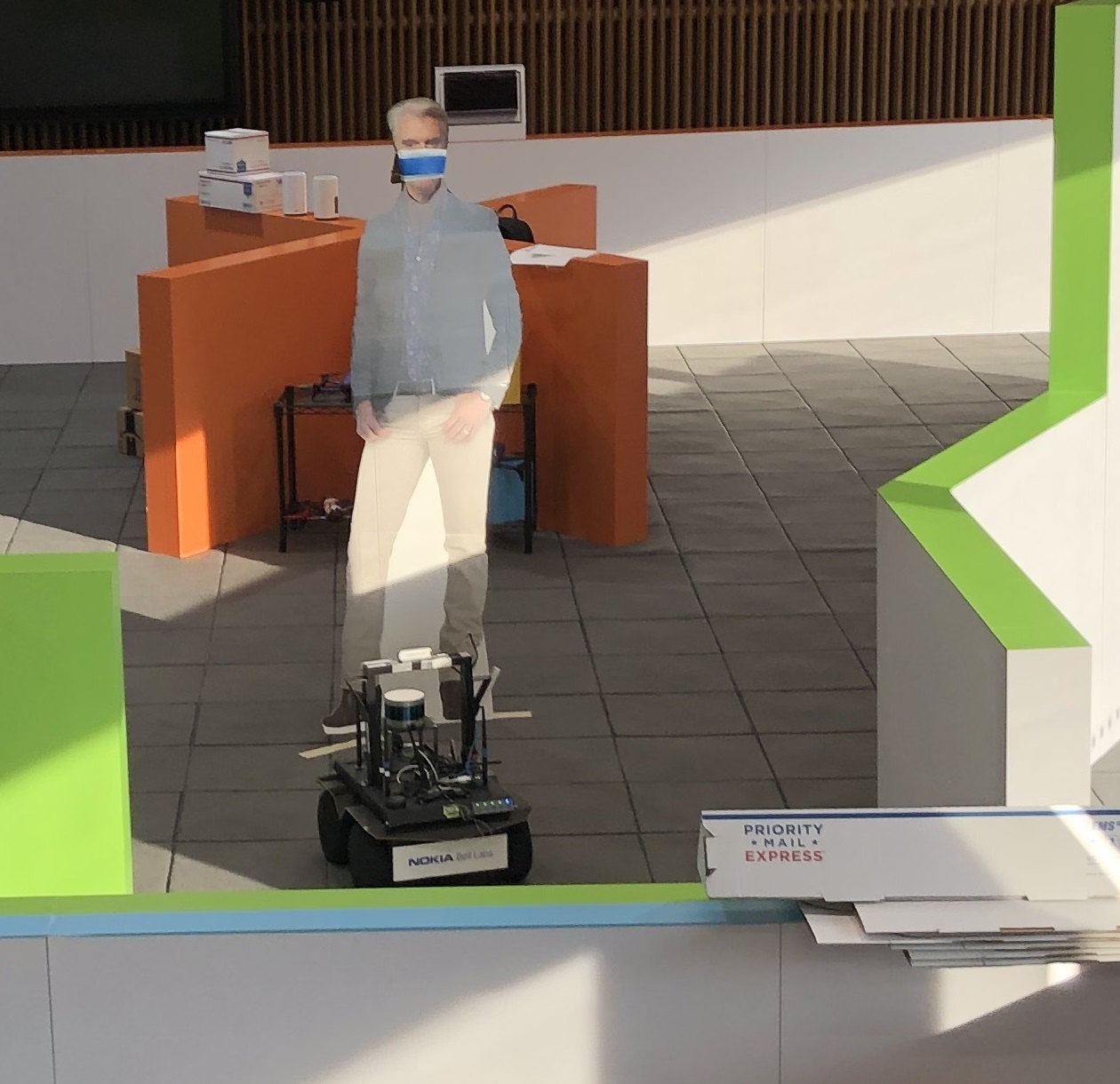}}\hspace{-1mm}	
	\subfigure{\label{fig:test3}
		\includegraphics[width=0.15\textwidth]{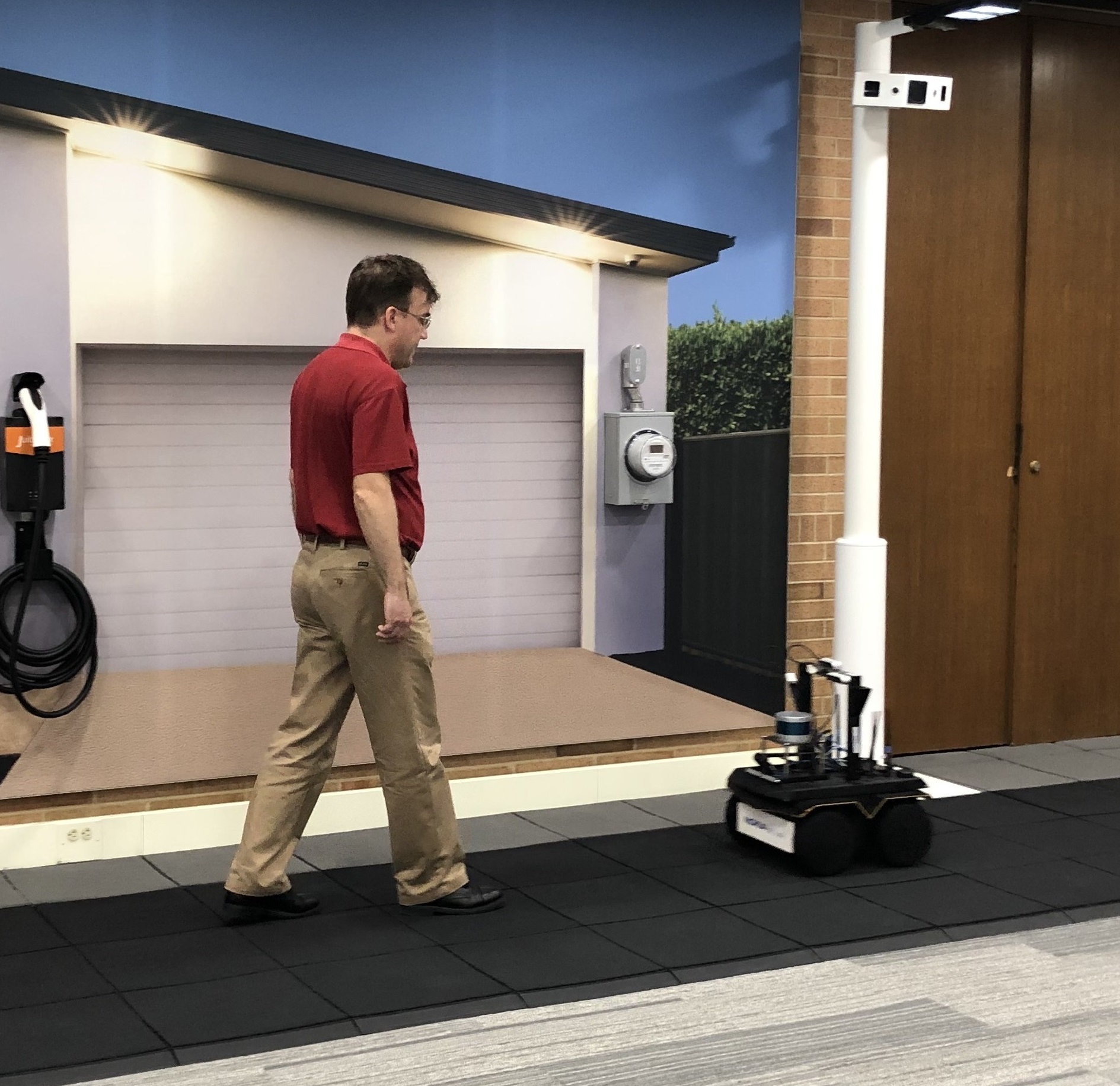}}\hspace{-1mm}	
	\vspace{-2mm}
	\caption{Robot experiment test cases.}\label{fig:test_cases}
\end{figure}

\noindent $\bullet$ \textbf{(C1) Room I to room N through doorway:} shown in Fig.\ref{fig:test_cases} (left). Here the robot's task is to travel through a narrow doorway while making a 180-degree turn. In this case, all the obstacles are fixed and included in the global map, and so the only job of the local planner is to follow the global plan (which will be a collision-free path from start to goal) as closely as possible. However, in order to make the turn smoothly the planner must maintain a small turn radius (the ratio between linear and angular velocity).\\
$\bullet$ \textbf{(C2) Room I to room X with ``unexpected'' static obstacle:} shown in Fig.\ref{fig:test_cases} (mid). In this experiment, the robot goal is selected before the obstacle is in place. After the goal is selected and the global plan is computed, a static obstacle (a cardboard cutout of a person) is placed in the robot's global plan. 
%The points on the global plan that are covered by the obstacle are not eligible to be waypoints.
As the robot nears the obstacle, the next eligible waypoint will be beyond the obstacle and the local planner will need to navigate round the obstacle.\\
$\bullet$ \textbf{(C3) Avoiding a walking pedestrian on a straight path:} shown in Fig.\ref{fig:test_cases} (right). Here the robot must traverse a straight path while a pedestrian is walking towards the robot. This case tests the local planner's ability to detect and navigate around a moving object. For this experiment it would always be possible to generate an ``unavoidable collision'' by having the pedestrian walk quickly at high speed into the robot. To avoid this we ask the pedestrian to stop when they are right in front of the robot. The desired behavior is then for the robot to back up or turn round the pedestrian. The undesired behavior is to keep on moving forward into the pedestrian.\\ 
\noindent $\bullet$ \textbf{(C4) Pedestrian crossing the robot path:} We extend the previous test case (C3) by asking the pedestrian to perpendicularly cross the robot's global path. The desired behavior is for the robot to wait and then continue after the pedestrian has crossed. 

\noindent \textbf{Local planners.} 
We test with the DrQ variant of SAC since it had the best training performance of all the RL algorithms in Section~\ref{sec:training}. We log the trajectories for the resulting {\em SACPlanner} and compare against the Dynamic Window Approach (DWA), as well as the Shortest Path (SP) planner discussed in Section~\ref{sec:previous} that always tries to get to the next waypoint using a shortest path in the Occupancy Grid. 
%DWA planner picks a candidate trajectory from the set of possible paths and abides to the picked trajectory's velocity vector. DWA planner persistently scans for the next best trajectory as the robot moves along the global path. 
%The shortest path planner uses the waypoint generator to cut the global path into several local paths. As the name implies, the goal is to reach the next waypoint by traveling the shortest distance between the current point and the next waypoint.
\iffalse
We test all planners on a Jackal robot fitted with a LiDAR scanner for sampling the current state.
As mentioned in the introduction, we upload the planners' modules to a central cloud computing node that contains ROS's navigation and local planning modules (along with the utility modules).
The planner communicates wirelessly with the robot's motors in order to execute the action at each timestep. 
For all experiments and planners, we set a timestep's duration to be $0.2$ seconds. In addition, we set the LiDAR's scanning frequency to be $5$ Hz and the local planner's inflation layer cost scaling factor and inflation radius to be $0.5$. 
For each of the planners' results, we show the global paths for $2$ runs given the same starting and goal points in the three experiments. For the static obstacle and walking pedestrian's cases, we note that the global plan is determined by the waypoint generator \emph{before} the pedestrian is detected by the local planner.
\fi 

%% file: experimentresults.tex
\section{Experimental Results}
\begin{figure*}[ht]
	\centering
	\subfigure{\label{fig:case1_comp}
		\includegraphics[width=0.24\textwidth]{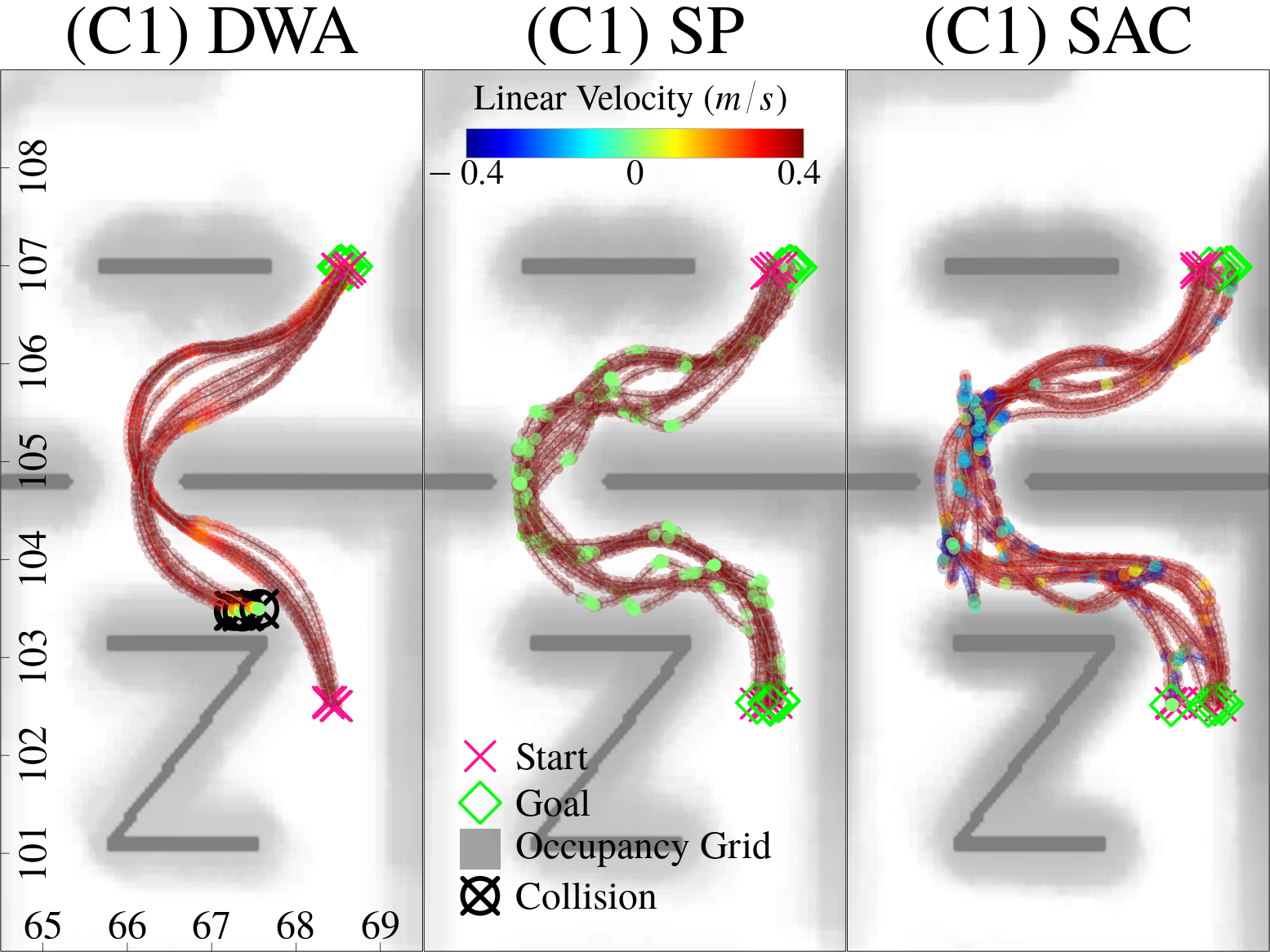}}\hspace{-1mm}
	\subfigure{\label{fig:case2_comp}
		\includegraphics[width=0.24\textwidth]{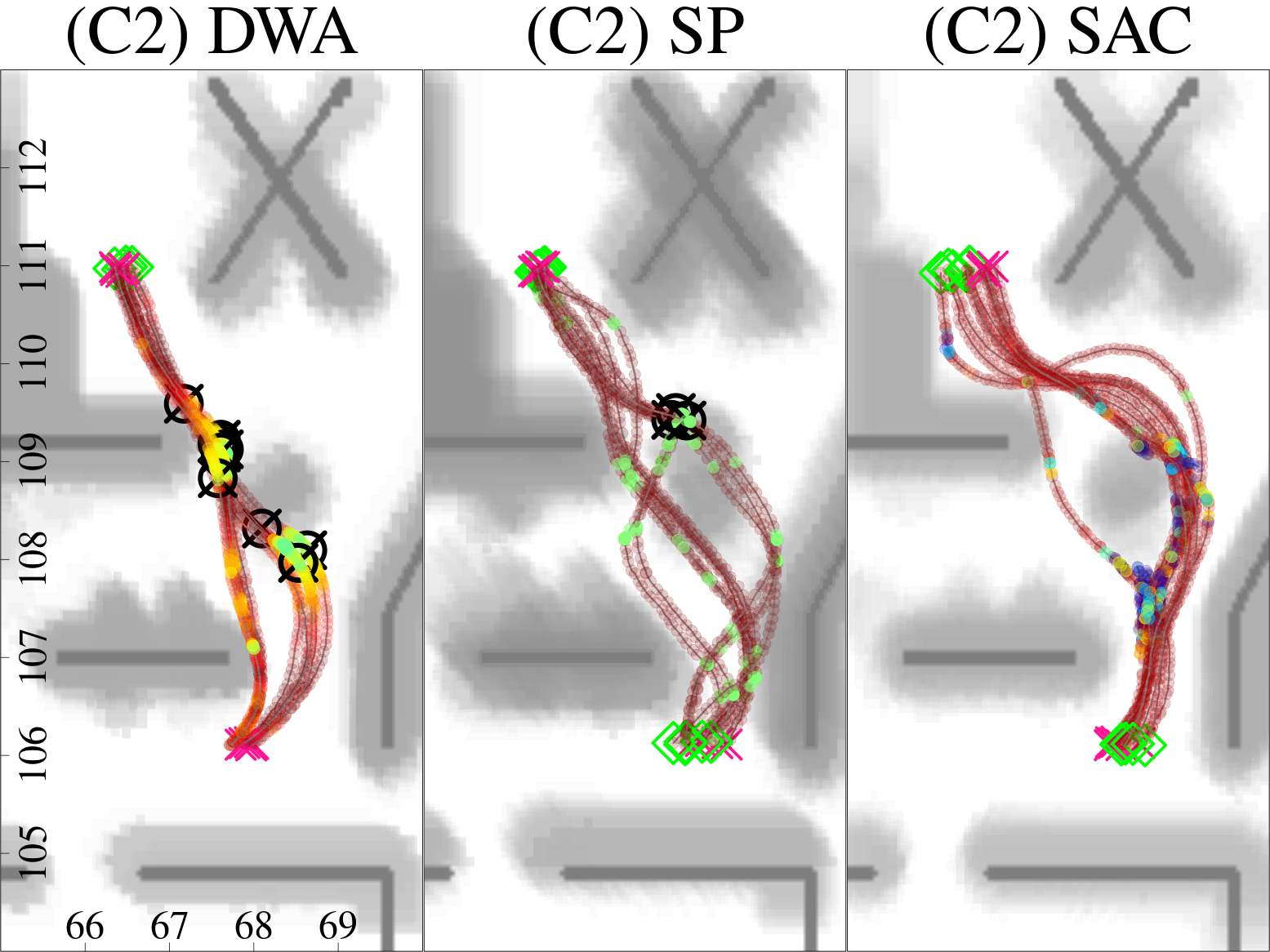}}\hspace{-1mm}	
	\subfigure{\label{fig:case3_comp}
		\includegraphics[width=0.24\textwidth]{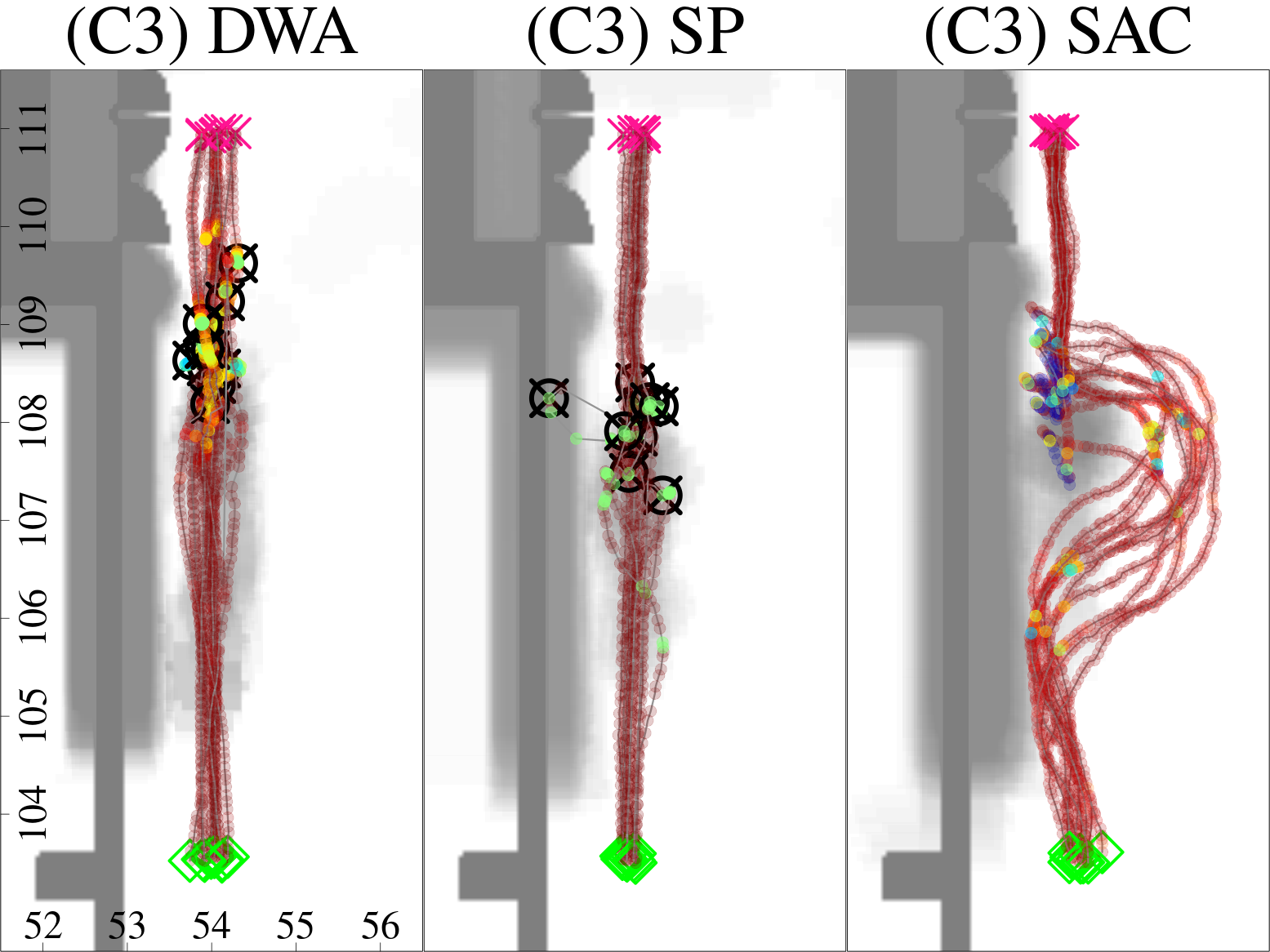}}\hspace{-1mm}	
	\subfigure{\label{fig:case4_comp}
		\includegraphics[width=0.24\textwidth]{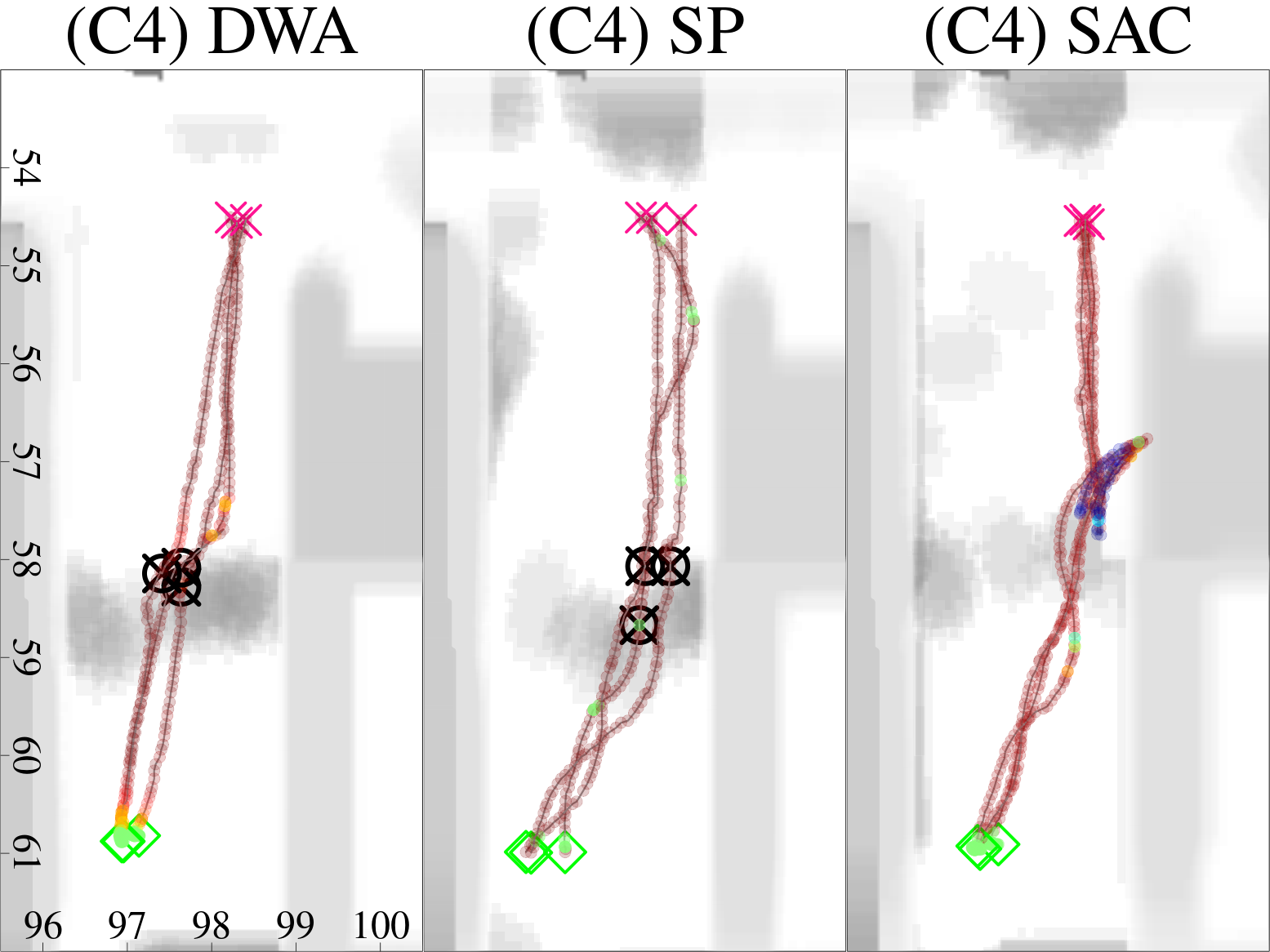}}	
	\vspace{-2mm}
	\caption{Trajectory comparison between DWA, Shortest Path (SP) vs. SAC agent for each test case.}\label{fig:case_comp}	
\end{figure*}

The robot trajectories for each of (C1)-(C4) are shown in Fig.\ref{fig:case_comp}. We denote the start and goal along with the collision points. For (C1)\&(C2) we swap the direction of travel for half the runs. The color of the trajectory represents linear velocity. We also show the Occupancy Grid values in gray (taken from the map and the LiDAR). For (C3)\&(C4) with a dynamic obstacle the gray shading captures all the positions of the obstacle over time. The 3 local planners have qualitatively different behavior which we now describe in detail for each case. 
%Considering the aforementioned characteristics of the movements in trajectory by each agent, we summarize here the performance of all $10$ robot runs driven by each agent using the DWA, Shortest Path (SP) and trained by the SAC algorithms for each test case (C1)-(C3) as in Fig.\ref{fig:case_comp} with the same color scheme. For (C1)\&(C2), we have switched the position between departure and goal for half times to investigate differences in left or right sharp turns in a narrow pathway.
\begin{figure*}[!th]
	\centering
	\subfigure{\label{fig:ped_traj}
	\includegraphics[width=0.22\textwidth]{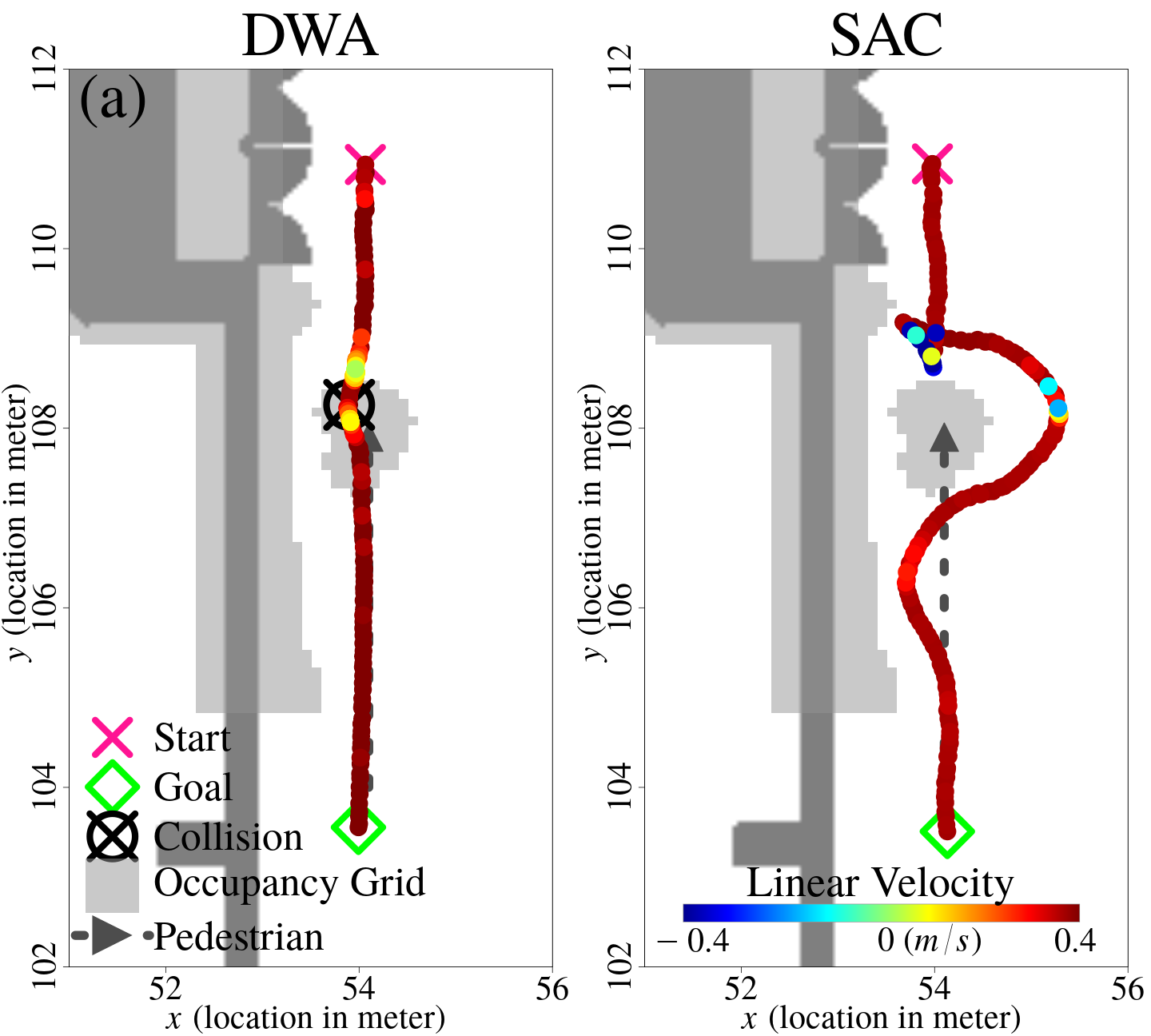}}	\hspace{-3mm}
	\subfigure{\label{fig:ped_linv}
	\includegraphics[width=0.22\textwidth]{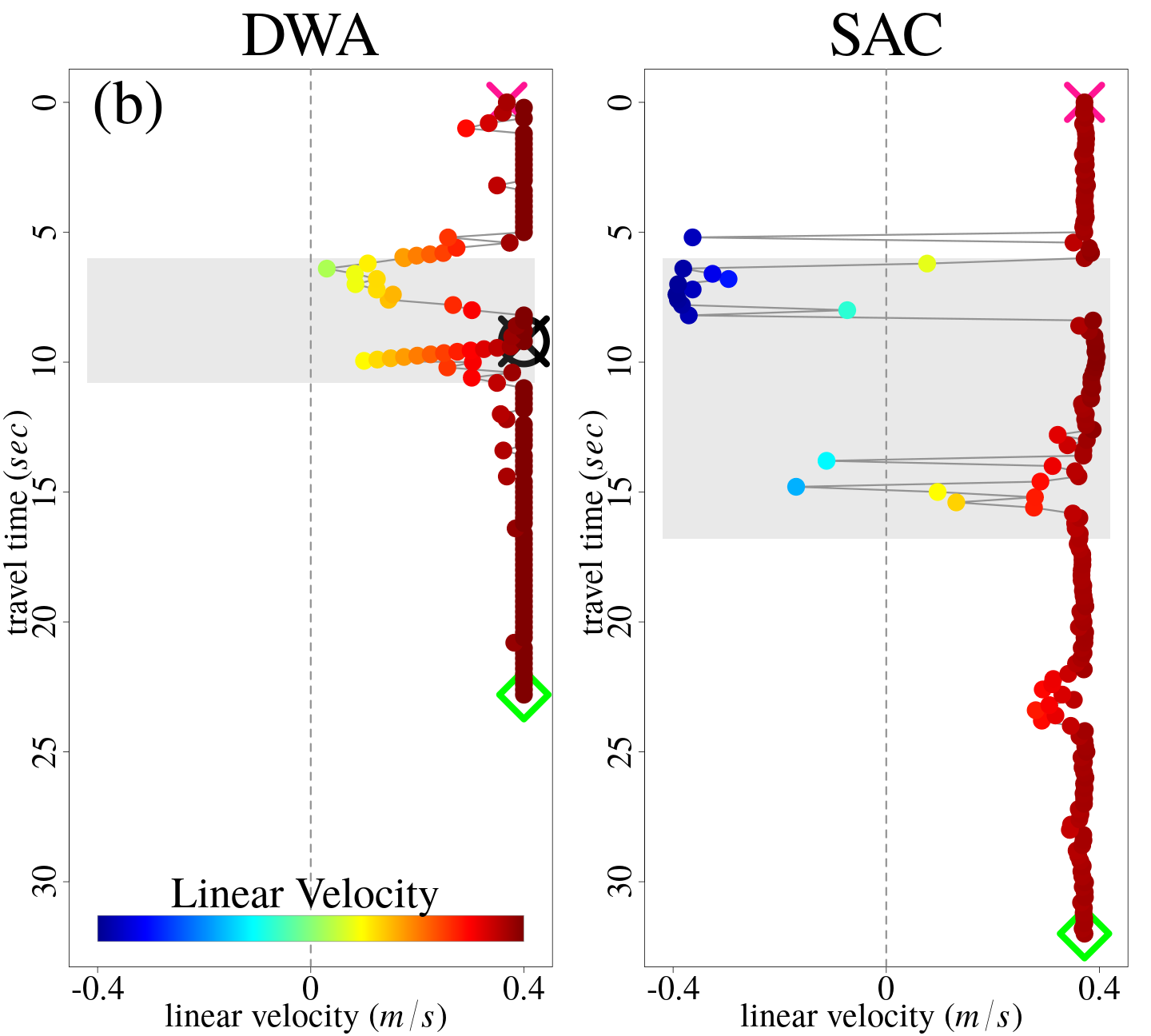}}	\hspace{-3mm}
	\subfigure{\label{fig:ped_mdist}
	\includegraphics[width=0.22\textwidth]{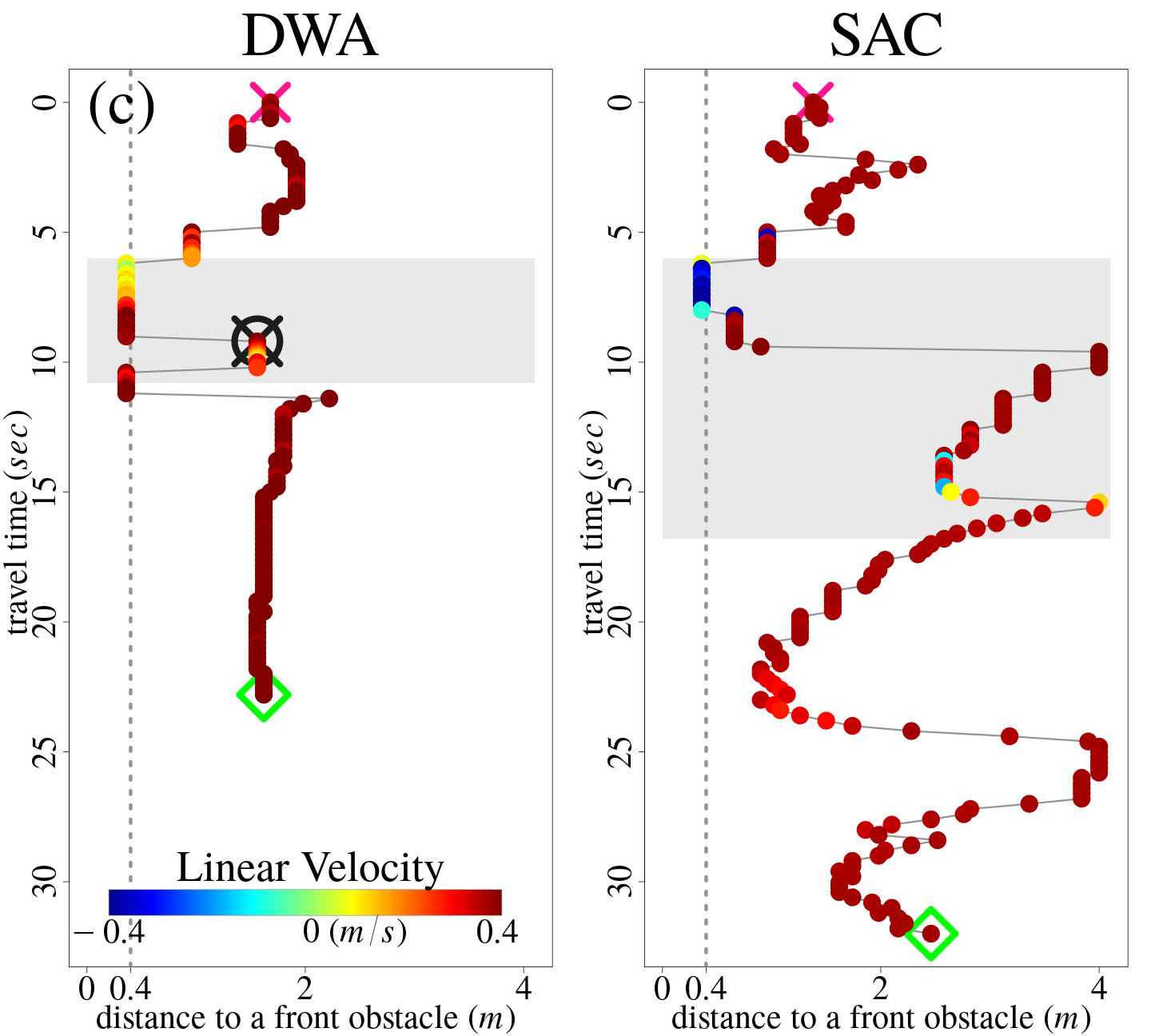}}	\hspace{-3mm}
	\subfigure{\label{fig:ped_angv}
	\includegraphics[width=0.22\textwidth]{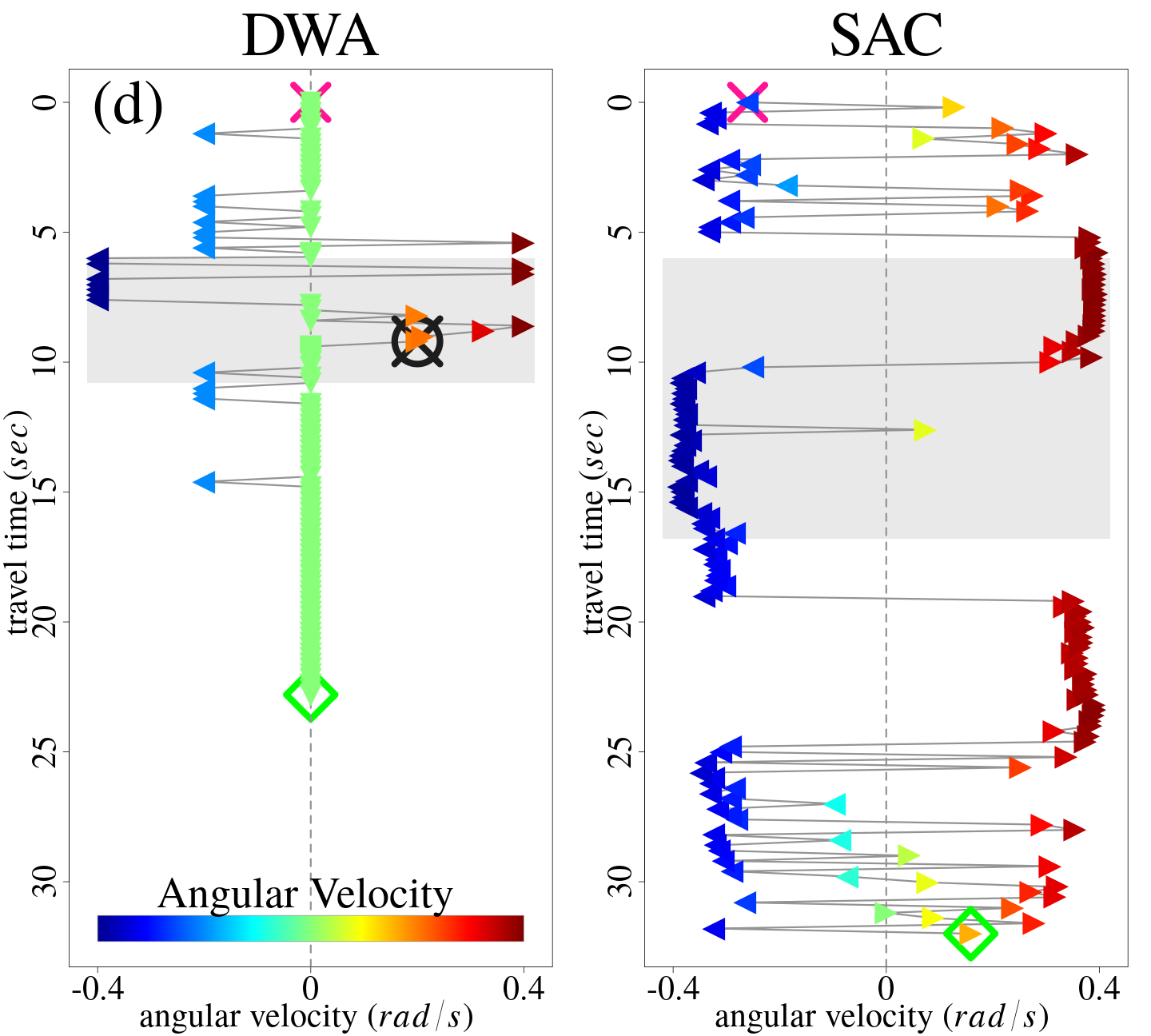}}	
\vspace{-3mm}
	\caption{Trajectory comparison between DWA and SACPlanner based on logs from the test case (C3).}\label{fig:ped_ex}
\end{figure*}

\noindent $\bullet$ \textbf{(C1):} DWA (which generates circular arcs) has the smoothest trajectory through the door. However, when starting at the top it miscalculated the best turning radius and aborted next to the `N' obstacle each time.
%collided to the wall in room N after a sharp left turn through the door due to the too fast speed for all $5$ times when it departed at upper $y$, and could not finish the whole journey by getting an abort control from the ROS. But it showed smooth trajectories in all the rest $5$ runs in the reverse traverse by getting to the goal successfully. 
The SP planner never collided with an obstacle and (not surprisingly since it was running shortest paths on a grid) it traveled in a series of straight lines (whose endpoints are denoted with green dots).  
%in all $10$ times, but it showed nearly `discrete' behavior selecting $3$ different types of actions in a sequence  `go-stop-turn' (like a machinery movement) as indicated by the straight lines between each two consecutive green dots in all trajectories.
SACPlanner was also successful in all cases. However, it had to ``back off'' multiple times  (denoted by the blue parts of the trajectory) before aligning correctly with the doorway. 
%got to the goal in all $10$ runs while almost equally balancing the both sides of the space between left and right through the narrow pathway, and it backed off (blue) at the door then go straight as in the vertical lines (like a cautious human movement).
   
\noindent $\bullet$ \textbf{(C2):} In this case a static obstacle appears on the global plan. Although DWA tried to deviate from the global plan, it did not do so enough, and therefore collided with the obstacle every time. 
%The DWA agent got too close to the walls and stepped on the bottom edge (the foot) of the cardboard obstacle in most of runs regardless of departure locations and even pushed it to be fell one occasion. In the meanwhile, when 
The SP planner was succesful when starting from the bottom. When starting from the top, the shortest path around the obstacle alternated between ``going left'' and ``going right''. This indecision led to some collisions. 
%agent departed from the upper $y$ position, the waypoint around the obstacle was pingponging between left and right due to the optimal route exploration since both are the shortest path around it, which caused the agent to push the cardboard while jiggling between both sides then even the robot rode over on top of the fallen obstacle in several times. 
SACPlanner often backed off multiple times when confronted with the obstacle. However, it eventually made it round the obstacle every time. %did back-and-forth multiple times which took more time to detour around the cardboard obstacle, but it all successfully got to the goal without any collision with overall smooth trajectories. 

\noindent $\bullet$ \textbf{(C3):} Both DWA and SP were unable to deal with the fact that the pedestrian obstacle was approaching and hence the ``correct'' trajectory kept changing. Even though the pedestrian stopped right in front of the robot, both DWA and SP kept going and caused a collision. SACPlanner went backwards when the pedestrian got close and then directed the robot to take a wide berth in the available open space. %agent detected the approaching obstacle earlier than the others and stopped for a few seconds, but eventually decided to force itself going forward and collided to the pedestrian in all $10$ runs. The SP agent reacted later or never than the other agents and collided either to the wall or the pedestrian every time. These are shown as the lower positions of collisions by the SP agent in terms of $y$-axis in the plot. The SAC agent successfully avoided the coming pedestrian by backing off then detoured widely enough by utilizing the open space and got to the goal by following the original path in the end for all $10$ runs. 

%In addition to these qualitative diagnosis of each agent's trajectories, 
\noindent \textbf{Quantitative metrics:} In Table \ref{tab:exp_results} we show the mean travel time ($s$), mean travel distance ($m$), mean speed ($m/s$), and collision rate for the 3 local planners on (C1)-(C3) across all runs. For DWA on (C1) we only consider the non-aborted runs. For (C2)-(C3) we remove the obstacle after each collision and so the robot will still reach the goal. We note that the ``backing off'' behavior of SACPlanner leads to greater distances/times than DWA and SP, but this how it is able to achieve a much lower collision rate. 

%(P4) distance to the nearest obstacle ($m$); and (P5) the standard deviation of `(P4)', (P6) count of collision, (P7) on average when the collision happened ($s$), and (P8) the mean yaw deviation (rad). Here (P3)$=$(P2)/(P1), and (P6) is the opposite of success rate since we removed obstacle after collision to let the agent finish its path except $5$ DWA (C1) with ROS abort. For (C1), the SP and SAC agents were the two best ones by balancing the space in the maze doorways with the good overall speed. For unexpected obstacle avoidance in (C2)\&(C3), the SAC agent was the best as explained before.   

\begin{table}[h!]
	\caption{Summary statistics of trajectories from test cases.}
	\vspace{-3mm}
	\label{tab:exp_results}
	\centering{\tiny%\scriptsize
	\begin{tabular}%{@{\hspace{1pt}}c@{\hspace{1pt}}|r|r|r|r|r|r|r|r|r}
	{c|r|r|r|r|r|r|r|r|r}
	\hline
& \multicolumn{3}{c|}{(C1)} & \multicolumn{3}{c|}{(C2)} & \multicolumn{3}{c}{(C3)} \\\hline
&	\multicolumn{1}{c|}{DWA}	&	\multicolumn{1}{c|}{SP}	&	\multicolumn{1}{c|}{SAC}	&	\multicolumn{1}{c|}{DWA}	&	\multicolumn{1}{c|}{SP}	&	\multicolumn{1}{c|}{SAC}		&	\multicolumn{1}{c|}{DWA}	&	\multicolumn{1}{c|}{SP}	&	\multicolumn{1}{c}{SAC}		\\\hline
Time	&	21.80	&	30.10	&	37.20	&	30.70	&	20.90	&	28.50	&	27.50	&	22.30	&	33.10	\\\hline
Distance	&	7.13	&	8.93	&	10.70	&	5.47	&	6.26	&	8.57	&	8.77	&	8.01	&	10.80	\\\hline
Speed	&	0.33	&	0.30	&	0.29	&	0.18	&	0.30	&	0.30	&	0.32	&	0.36	&	0.33	\\\hline
%mean obs dist	&	0.08	&	{\bf 0.22}	&	{\bf 0.22}	&	0.16	&	0.21	&	{\bf 0.30}	&	0.28	&	0.24	&	{\bf 0.31}	\\\hline
%stddev obs dist	&	0.04	&	0.03	&	0.05	&	0.07	&	0.04	&	0.06	&	0.25	&	0.12	&	0.08	\\\hline
Collision	&	0.5	&	{\bf 0}	&	{\bf 0}	&	1.0	&	0.3	&	{\bf 0}	&	1.0	&	0.9	&	{\bf 0}	\\\hline
%collision time	&	15.90	&		&		&	9.46	&	6.40	&		&	7.88	&	8.17	&		\\\hline
%yaw dev	&	0.26	&	0.36	&	0.33	&	0.20	&	0.34	&	0.31	&	0.17	&	0.22	&	0.33	\\\hline
	\end{tabular}}
\end{table}

\noindent $\bullet$ \textbf{(C4)} When the pedestrian switches to walking across the robot's path rather than walking towards it, the results are similar to (C3). 
%setting the pedestrian to perpendicularly cut cross the robot's global path in a wider open space to investigate each agent's reaction. Here the global costmap does not contain any static obstacle, but the chairs and tables around this area appears in the local costmap at the left side of robot's traverse direction. We did $3$ runs each with all the agents in this case, and the trajectories are also displayed in Fig.\ref{fig:case4_comp} where the gray blobs smudged as horizontally reflecting the dynamic movement of the crossing pedestrian. 
Both DWA and SP are not reactive enough and collide every time. However, SACPlanner backs off when the pedestrian is close, and then resumes traveling towards the goal after the pedestrian has passed through.   

%% file: trajectory.tex
\subsection{Trajectory Analysis}\label{sec:trajectory}
In order to understand more deeply the difference in behavior of DWA and SACPlanner,  Fig.\ref{fig:ped_ex} depicts a single run from test case (C3). 
%We analyze the trajectories from log data to understand how the velocity action commands translate into the actual movements by the test case (C3) pictured in Fig.\ref{fig:test3} run by the DWA and SAC agents as in Fig.\ref{fig:ped_ex}. 
Fig.\ref{fig:ped_traj} shows the trajectory, Fig.\ref{fig:ped_linv} plots the linear velocity, Fig.\ref{fig:ped_mdist} shows the distance to the nearest `front obstacle' (within $\pm\frac{\pi}{4}$rad range from the current yaw), and Fig.\ref{fig:ped_angv} plots the angular velocity. 

The key feature of these plots is that when the pedestrian is close, DWA slows down and turns a little, whereas SACPlanner goes into reverse (note the blue color in Fig.\ref{fig:ped_linv}\&(c)) and turns a lot so as to go around the pedestrian. This ``reactiveness'' to obstacles also manifests in more turning even when the robot can go in a straight line.